%% file: main.tex
\documentclass{article}
\pdfoutput=1
\usepackage[square, comma, sort&compress, numbers]{natbib}

\usepackage[preprint]{neurips_2024}

\usepackage[utf8]{inputenc} 
\usepackage[T1]{fontenc}    
\usepackage{url}            
\usepackage{booktabs}       
\usepackage{amsfonts}       
\usepackage{nicefrac}       
\usepackage{microtype}      
\usepackage{xcolor}         

\definecolor{mycitecolor}{rgb}{0, 0.4, 0.7}
\usepackage[hidelinks,colorlinks=true,citecolor=mycitecolor]{hyperref}
\usepackage{graphicx} 
\usepackage{multirow}
\usepackage{amsmath}
\newcommand*\rot{\rotatebox{90}}

\title{Multi-Modal Generative Embedding Model}

\author{%
Feipeng Ma$^{1}$\thanks{This work was performed while Feipeng Ma and Hongwei Xue were interns at WeChat, Tencent Inc.}, Hongwei Xue$^{1,3}$\thanks{Project Leader.}, Guangting Wang$^{2}$, Yizhou Zhou$^{2}$\footnotemark[3], Fengyun Rao$^{2}$ \\ {\bf Shilin Yan$^{4}$, Yueyi Zhang$^{1}$, Siying Wu$^{5}$, Mike Zheng Shou$^{3}$, Xiaoyan Sun$^{1,5}$\thanks{Corresponding authors.} }\\
  \\
  $^{1}$University of Science and Technology of China \quad $^{2}$WeChat, Tencent Inc. \qquad \\
  $^{3}$Show Lab, National University of Singapore \qquad $^{4}$ Fudan University \\
  $^{5}$Institute of Artificial Intelligence, Hefei Comprehensive National Science Center
  \\
  \texttt{\{mafp,xuehongwei\}@mail.ustc.edu.cn} \\
 \texttt{harryizzhou@tencent.com},
  \texttt{sunxiaoyan@ustc.edu.cn} \\
}

\begin{document}

\maketitle

\input{sections/00_abstract}
\input{sections/01_intro}
\input{sections/02_related}
\input{sections/03_method}

\input{sections/04_experiment}

\input{sections/05_conclusion}


{\small
\bibliographystyle{plain}
\bibliography{main}
}


\end{document}

%% file: sections/00_abstract.tex
\begin{abstract}
  Most multi-modal tasks can be formulated into problems of either generation or embedding. Existing models usually tackle these two types of problems by decoupling language modules into a text decoder for generation, and a text encoder for embedding. To explore the minimalism of multi-modal paradigms, we attempt to achieve \textit{only one model per modality} in this work. We propose a Multi-Modal Generative Embedding Model (MM-GEM), whereby the generative and embedding objectives are encapsulated in one Large Language Model. We also propose a PoolAggregator to boost efficiency and enable the ability of fine-grained embedding and generation. A surprising finding is that these two objectives do not significantly conflict with each other. For example, MM-GEM instantiated from ViT-Large and TinyLlama shows competitive performance on benchmarks for multimodal embedding models such as cross-modal retrieval and zero-shot classification, while has good ability of image captioning. Additionally, MM-GEM can seamlessly execute region-level image caption generation and retrieval tasks. Besides, the advanced text model in MM-GEM brings over 5\% improvement in Recall@1 for long text and image retrieval. 
\end{abstract}

%% file: sections/01_intro.tex
\section{Introduction}\label{sec:intro}
In recent years, the multi-modal learning field has witnessed a unifying trend \cite{li2022blip,bao2022vlmo,lu2022unified,zhou2020unified,huang2021unifying}. This trend is driven by the advanced understanding ability and more efficient computation brought by shared representations. Moreover, the simplification of the model structure makes it much more direct to perform various downstream tasks \cite{radford2021learning,li2022blip}.

Most cutting-edge multi-modal models can be categorized into two paradigms: embedding models and generative models. Embedding models \cite{radford2021learning,jia2021scaling,xue2022clip} typically utilize a dual encoder structure. This framework projects distinct modalities into a unified latent space, thereby facilitating efficient cross-modal retrieval and classification tasks. Generative models \cite{zhu2023minigpt,liu2024visual,alayrac2022flamingo} forge a connection between visual representations and Large Language Models (LLMs). This integration enables the models to harness capabilities such as instruction following \cite{zhu2023minigpt,liu2024visual} or in-context learning \cite{alayrac2022flamingo}. These two paradigms intersect in the visual modality, i.e., the vision module of the generative model is usually derived from a powerful embedding model \cite{radford2021learning}. However, the textual modality reveals a divergence in approach. While generative models commonly employ an auto-regressive text decoder, embedding models favor a text encoder to extract a global representation of the text.

The divergence in textual modality remains as a pivotal obstacle to achieving the goal of unification, namely, using \textit{only one model per modality}. several works step in different directions towards this goal. BLIP \cite{li2022blip} shares all parameters in the text encoder and decoder except for the self-attention layers. CoCa \cite{yu2022coca} splits the text decoder into unimodal and multimodal components, by removing the cross-attention module in the unimodal decoder layers. These methods differentiate the forward path of unimodal and multimodal, introducing a hindrance to the direct use of pre-trained text models. FROMAGe \cite{koh2023grounding} truly achieves unification by grounding the image feature to the inputs and outputs of a frozen large language model. However, the lack of joint training with the visual modality results in a performance deficiency.

To explore the minimalism of multi-modal paradigms, we propose Multi-Modal Generative Embedding Model (MM-GEM) in this paper. MM-GEM is an end-to-end optimized model that combines two paradigms by encapsulating the generative and embedding objectives in the same language model. Specifically, for embedding, we align the image features with the sentence embeddings derived from the last token. Concurrently, for the generative task, we concatenate the image features with the word embeddings of the language model to execute the captioning process. Notably, both objectives leverage a shared forward path within the language model. To boost efficiency and enable the ability of fine-grained embedding and generation, we propose a PoolAggregator to represent an image by the feature map, instead of a global feature on [CLS] token \cite{radford2021learning,dosovitskiy2020image}. 

Experimental results demonstrate the superiority of MM-GEM. MM-GEM instantiated from ViT-Large and TinyLlama \cite{zhang2024tinyllama} achieves comparable results with OpenCLIP \cite{openclip} on image-text retrieval benchmarks such as COCO \cite{lin2014microsoft} and Flickr30K \cite{plummer2015flickr30k}, and zero-shot image classification benchmark ICinW \cite{li2022elevater}. Meanwhile, MM-GEM shows competitive performance on image captioning benchmarks such as COCO \cite{lin2014microsoft} and NoCaps \cite{agrawal2019nocaps}. Additionally, qualitative results show that MM-GEM can generate region-level image captions and fine-grained text-to-image retrieval without further training or modification. Besides, the advanced text module in MM-GEM brings better ability of text understanding. MM-GEM achieves over 5\% higher Recall@1 for long text and image retrieval, compared to CLIP. 

Our contributions are summarized as follows:
\begin{enumerate}
\item We propose a Multi-Modal Generative Embedding Model (MM-GEM), whereby the generative and embedding objectives are encapsulated to achieve unification.
\item A PoolAggregator and Multi-Stage Training strategy are proposed to represent an image from the feature map, which efficiently enables the fine-grained ability.
\item Experimental result demonstrates that MM-GEM shows competitive performance on benchmarks for embedding models, while still keeps the good ability of generation.
\end{enumerate}

%% file: sections/02_related.tex
\section{Related Works}
\textbf{Vision-Language Pre-trained Models} have undergone a significant evolution in their application to downstream tasks. Earlier works \cite{chen2020uniter,huang2020pixel,kim2021vilt,xue2021probing,xue2022advancing,lei2021less} learn cross-modal representations through proxy tasks such as masked modeling and image-text matching, but require further fine-tuning for specific downstream tasks. Modern works build pre-trained models in a manner that mirrors their ultimate use to achieve seamless integration with downstream tasks. These works can be categorized into two paradigms: embedding models and generative models. Embedding models \cite{radford2021learning,jia2021scaling,xue2022clip} independently extract and map features from each modality into a shared space, facilitating efficient cross-modal retrieval and open-set classification. Generative models \cite{wang2022git,zhu2023minigpt,liu2024visual,alayrac2022flamingo} reformulate downstream tasks like Visual Question Answering (VQA) as auto-regressive generation tasks. Inheriting capabilities such as instruction following \cite{zhu2023minigpt,liu2024visual} or in-context learning \cite{alayrac2022flamingo} from LLMs, generative models are often utilized to tackle complex understanding tasks that cannot be well solved by embedding models.

\textbf{Modality Unification} has been a long standing goal to pursue better understanding ability and higher computation efficiency brought by shared representations. However, the distinct requirements of embedding and generative models often lead to the preference for separate text encoders and decoders. BLIP \cite{li2022blip} and InternVL \cite{chen2023internvl} approach this by sharing most parameters across the text encoder and decoder, with the exception of the self-attention or cross-attention layers. CoCa \cite{yu2022coca} splits the text decoder into unimodal and multimodal components, then remove the cross-attention module in the unimodal decoder layers. FROMAGe \cite{koh2023grounding} grounds the image feature to the inputs and outputs of a frozen large language model by several projection layers. However, the lack of joint training results in a performance deficiency. One very recent work GRIT \cite{muennighoff2024generative} successfully unifies embedding and generative Natural Language Processing (NLP) tasks. It is worth noting that unimodal models imply a natural correspondence in embedding. For example, generative-only models show a certain level of performance on embedding NLP tasks \cite{muennighoff2024generative}. However, whether there is a significant conflict between the multi-modal embedding and generative objectives remains under explored. 

%% file: sections/03_method.tex
\begin{figure}[t]
  \centering
  \includegraphics[width=\textwidth]{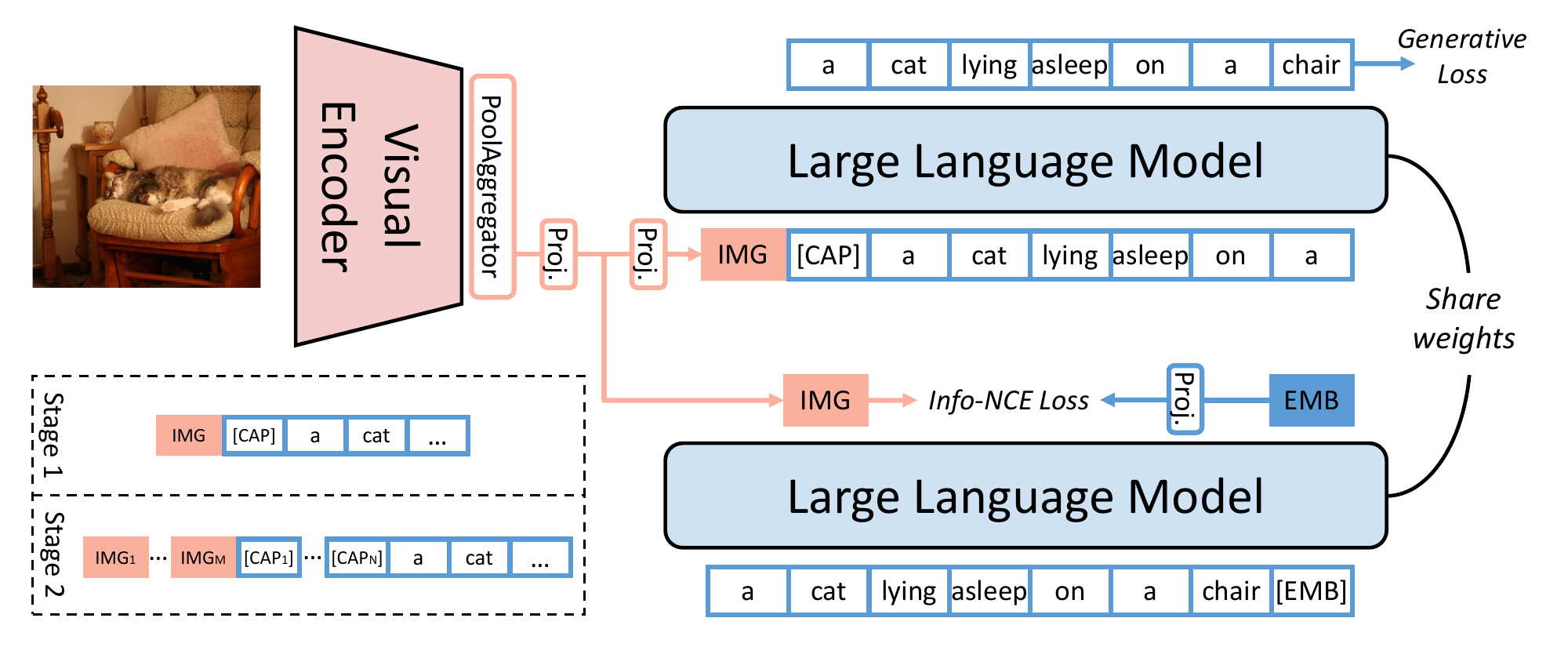}
  \caption{Overview of MM-GEM, in which a large language model acts as both text encoder for embedding and text decoder for generation. The visual feature is aligned with the LLM by several projection layers and a PoolAggregator.}
\end{figure}

\section{Approach}
The primary goal of Multi-Modal Generative Embedding Model (MM-GEM) is unifying text encoder and decoder, while supporting both embedding and generative paradigms. However, there are three uncertainties or challenges to achieve this unification: 1) Whether the superposition of two learning targets in the same language model will significantly conflict with each other. 2) Embedding models like CLIP extract one global visual feature for each image, which is insufficient for generative tasks. 3) The data suitable for training embedding models is often noisy, which resulting in the sub-optimal ability of generation. In this section, we will introduce our approach of tackling these three problems.

\subsection{Encapsulated Generative and Embedding Objectives}
A generative language model consists of a word embedding layer, a stacked transformer, and a prediction head. The word embedding layer can be regarded as a linear transformation to project the one-hot probability distribution $p_{i=1,2,..,L}$ of the input sequence to the latent space $\mathcal{W}_{\text{in}}$. The prediction head acts as a linear transformation to project the latent space $\mathcal{W}_{\text{out}}$ to the unnormalized probability distribution $\hat{p}_{i=1,2,..,L}$. As the $\hat{p}_t$ is a well estimation of $p_{t+1}$, the transformation between latent space $\mathcal{W}_{\text{in}}$ and $\mathcal{W}_{\text{out}}$ is approximately linear. Therefore, we leverage light projection layers $h_1$, $h_2$ and $h_3$ to transform latent space between visual space $\mathcal{V}$, $\mathcal{W}_{\text{in}}$ and $\mathcal{W}_{\text{out}}$:
\begin{equation}
    \mathcal{V}_{\text{Emb}} = h_1(\mathcal{V}),~~~~ \mathcal{W}_{\text{Emb}} = h_2(\mathcal{W}_{\text{out}}),~~~~ \mathcal{V}_{\text{in}} = h_3(\mathcal{V}_{\text{Emb}}),
\end{equation}
where $\mathcal{V}_{\text{Emb}}$ and $\mathcal{W}_{\text{Emb}}$ is the space of image and text embeddings, respectively. $\mathcal{V}_{\text{in}}$ is the space of visual features $V_{\text{in}}$ to be concatenated with word embeddings from space $\mathcal{W}_{\text{in}}$.

For embedding, we follow \cite{radford2021learning} and adopt the info-NCE loss to learning the cross-modal alignment:
\begin{equation}
\mathcal{L}_{\text{v2t}} =-\frac{1}{B} \sum_{i=1}^{B} \log \frac{e^{v_{i}^{\top} t_{i} / \tau}}{\sum_{j=1}^{B} e^{v_{i}^{\top} t_{j} / \tau}}, ~~~~
\mathcal{L}_{\text{t2v}} =-\frac{1}{B} \sum_{i=1}^{B} \log \frac{e^{t_{i}^{\top} v_{i} / \tau}}{\sum_{j=1}^{B} e^{t_{i}^{\top} v_{j} / \tau}},
\end{equation}
and $\mathcal{L}_{\text{Emb}} = \mathcal{L}_{\text{v2t}} + \mathcal{L}_{\text{t2v}}$, where $v_i$ and $t_j$ are the normalized embeddings of $i$-th visual feature and $j$-th text feature in a batch of size $B$. $\tau$ is a learnable temperature. The visual feature will be discussed in Section \ref{sec:3.2} and the text feature is the last hidden states on top of a [EMB] token appended to the text sequence.

For generation, we adopt the image captioning loss to predict the next token $x^{(i)}$ based on the visual input $V_{\text{in}}$, a special [CAP] token, and previous tokens $x^{(<i)}$:
\begin{equation}
    \mathcal{L}_{\text{Gen}}=-\frac{1}{B}\sum_{i=1}^B \log P(f_{\theta}(x^{(i)}) | f_{\theta}([V_{\text{in}}, \text{[CAP]}, x^{(<i)}])),
\end{equation}
where $f_{\theta}$ is the language model parameterized by $\theta$ and [*] is the concatenate operation. 

During training, both learning objectives are applied on all data samples. The language model forwards twice to get two losses and the final loss function is the direct summation without careful adjustment of weighting factors:
\begin{equation}
    \mathcal{L}_{\text{MM-GEM}} = \mathcal{L}_{\text{Emb}} + \mathcal{L}_{\text{Gen}}.
\end{equation}

\subsection{Vision PoolAggregator and Multi-Stage Training}\label{sec:3.2}
Simultaneously training generative and embedding objectives encounters two main challenges: 1) embedding models usually take the global feature as visual representation, while a global feature is insufficient for image caption generation. 2) the data suitable for training CLIP is large-scale alt-texts. Sometimes a alt-text is noisy, lacking of linguistic coherence, and the relevance to visual information is weak. To alleviate these issues, we propose a multi-stage training strategy equipped with a Vision PoolAggregator.

Instead of extracting a global visual feature on top of a [CLS] token, MM-GEM represents an image as a spatial feature map $V \in \mathbb{R}^{C \times H \times W}$, where $C$, $H$ and $W$ indicate the dimension, height and width of the feature map. A Vision PoolAggregator flexibly aggregates the visual information from $V$. To boost the training efficiency, in stage-one pre-training, we apply both embedding and generative loss on the mean-pooled visual feature:
\begin{equation}
    V_{\text{in}}^{1} = h_3(h_1(\text{MeanPool}(V))),~~~~V_{\text{Emb}} = h_1(\text{MeanPool}(V)).
\end{equation}
This significantly improves the training efficiency of stage one, and enables larger batch size preferred by contrastive learning. To further improve MM-GEM's ability of generation and fine-grained understanding, we set a stage-two pre-training procedure on image-caption and dense caption data. In this stage, visual feature $V$ is aggregated to a fix-sized feature map shape of $H \times W$ according to the region $R$ of the description (whole image for regular image-caption data):
\begin{equation}
    V_{\text{in}}^{2} = h_3(h_1(\text{RoIAlign}(V, R))).
\end{equation}
In stage two, the [CAP] token is replaced with a series of soft prompts $[\text{CAP}_{i}], i \in [1,N]$, only $h_3$ layer and soft prompts are updated in this stage. Therefore, the embedding ability learned in stage one is completely maintained. In parallel with dense captioning, we also boost MM-GEM's fine-grained retrieval ability by training a head on regional image-text pairs. To eliminate interference with existing abilities, we add a linear head $h_4$ on the output of $h_1$:
\begin{equation}
    V_{\text{Emb}}^2 = h_4(h_1(\text{RoIAlign}(V, R))),
\end{equation}
where RoIAlign aggregates visual feature $V$ to embedding according to the region $R$. We find that original CLIP or Captioning model fails in focusing regional visual information, which is dramatically improved by the proposed stage two. More details will be discussed in Section \ref{sec:finegrain}.

%% file: sections/04_experiment.tex
\section{Experiments}

\subsection{Implementation Details}
\noindent\textbf{Pre-training Stage One.}
We pre-train MM-GEM using LAION-2B~\cite{schuhmann2022laion} and COYO-700M~\cite{kakaobrain2022coyo-700m} datasets, containing a total of 2.3 billion image-text pairs. The language model is initialized from   TinyLlama~\cite{zhang2024tinyllama}. For the visual encoder, we study three variants of ViTs: ViT-B/16, ViT-L/14, and ViT-L/14-336, and these visual encoders are initialized from OpenCLIP \cite{openclip}. We use LAMB~\cite{you2019large} optimizer with a weight decay of 0.05. 
The learning rate for the projection layers is set to 5$e$-4, while for the visual encoder and the large language model, it is set to 5$e$-5. Input images are randomly cropped to a resolution of 224$\times$224 during pre-training except for ViT-L/14-336. We adopt a linear warm-up then cosine decay learning rate schedule. The training procedure is performed on a total batch size of 81,920 for 80,000 iterations.
The text processing follows TinyLlama except that the maximum length of the text is truncated to 50. Following CLIP \cite{radford2021learning}, the learnable temperature parameter $\tau$ is initialized to 0.07 and clipped at 0.01. We use 64 $\times$ H800 GPUs to train the model in this stage.

\begin{table}[t]
  \caption{Performance comparison on COCO \cite{lin2014microsoft} and Flickr30K \cite{plummer2015flickr30k} image-to-text (I2T) and text-to-image (T2I) retrieval. R@1, R@5, and R@10 indicate recall ratio at top 1, 5, and 10. All models in this table use ViT-Base as vision encoder.}
  \label{tab:base_retrieval}
  \centering
  \resizebox{\linewidth}{!}{
  \begin{tabular}{l ccc ccc ccc ccc}
    \toprule
    \multirow{2}{*}{Model} & \multicolumn{3}{c}{COCO I2T} & \multicolumn{3}{c}{COCO T2I} & \multicolumn{3}{c}{Flickr30K I2T} & \multicolumn{3}{c}{Flickr30K T2I} \\
    \cmidrule(lr){2-4}\cmidrule(lr){5-7}\cmidrule(lr){8-10}\cmidrule(lr){11-13}
    & R@1  & R@5  & R@10  & R@1  & R@5  & R@10  & R@1  & R@5  & R@10  & R@1  & R@5  & R@10  \\
    \midrule
    CLIP-Only & 57.8 & 80.3 & 87.3 & 41.7 & 67.2 & 76.7 & 86.2 & 97.7 & 99.2 & 70.3 & 91.4 & 95.3    \\
    MM-GEM & 57.0 & 79.7 & 87.2 & 41.4 & 66.6 & 76.3 & 84.6 & 97.5 & 99.3 & 70.1 & 90.7 & 95.0 \\
    
    \bottomrule
  \end{tabular}
  }
\end{table}

\begin{table}[t]
  \setlength\tabcolsep{2pt}
  \caption{Performance comparison on Image Classification in the Wild (ICinW) \cite{li2022elevater}. The metric of each dataset follows \cite{li2022elevater}. All models in this table use ViT-Base as vision encoder.}
  \label{tab:base_cls}
  \centering
  \resizebox{\linewidth}{!}{
  \begin{tabular}{l ccccc ccccc ccccc ccccc c}
    \toprule
    Model & \rot{\tiny{CIFAR-10~\cite{cifar}}} & \rot{\tiny{CIFAR-100~\cite{cifar}}} & \rot{\tiny{DTD~\cite{dtd}}} & \rot{\tiny{EuroSAT~\cite{eurosat}}} & \rot{\tiny{FER-2013~\cite{radford2021learning}}} & \rot{\tiny{FGVC-Aircraft~\cite{fgvc}}} & \rot{\tiny{KITTI-Dist.~\cite{kitti}}} & \rot{\tiny{MNIST~\cite{mnist}}} & \rot{\tiny{PatchCamelyon~\cite{patchcamelyon}}} & \rot{\tiny{VOC-2007~\cite{voc}}} & \rot{\tiny{Caltech-101~\cite{caltech}}} & \rot{\tiny{Country-211~\cite{radford2021learning}}} &	\rot{\tiny{Food-101~\cite{food101}}} & \rot{\tiny{GTSRB~\cite{gtsrb}}} &	\rot{\tiny{Hateful Memes~\cite{hateful}}} & \rot{\tiny{Ox. Flowers~\cite{flowers}}} & \rot{\tiny{Ox. IIIT Pets~\cite{pets}}} & \rot{\tiny{Rendered-SST2~\cite{radford2021learning}}} &	\rot{\tiny{RESISC-45~\cite{resisc45}}} &	\rot{\tiny{Stanford-Cars~\cite{stanford_cars}}} & \rot{\tiny{Average}}  \\
    \midrule
    CLIP-Only   & 94.8 & 75.5 & 59.2 & 54.7 & 38.6 & 20.2 & 27.0 & 70.3 & 50.1 & 79.9 & 90.3 & 18.6 & 82.3 & 42.4 & 57.7 & 58.1 & 84.0 & 59.9 & 64.6 & 81.8 & 60.5   \\
    MM-GEM   & 94.6 & 75.8 & 60.5 & 54.8 & 43.7 & 19.9 & 12.1 & 73.8 & 50.4 & 80.2 & 90.2 & 18.2 & 82.1 & 48.1 & 58.4 & 57.6 & 81.0 & 63.8 & 66.6 & 80.5 & 60.6   \\
    \bottomrule
  \end{tabular}
  }
\end{table}

\noindent\textbf{Pre-training Stage Two.}
In stage two, For fine-grained captioning, MM-GEM is further trained on CC3M \cite{sharma2018conceptual}, CC12M \cite{changpinyo2021conceptual}, SBU \cite{ordonez2011im2text} and LAION \cite{schuhmann2021laion} filtered by BLIP \cite{li2022blip} and Visual Genome's dense caption data \cite{krishna2017visual}, containing 38 million image-caption and 1.8 million region-description pairs in total. In this stage, only the $h_3$ layer and soft prompts are updated. The number of soft prompts is set as 64. This training procedure utilizes a total batch size of 2048 for 60,000 iterations, and the learning rate is consistent with that of stage one.

For fine-grained retrieval, the training data is in line with fine-grained captioning. The training procedure is performed on a total batch size of 49,152 for 15,000 iterations. To avoid disrupting the alignment learned in Stage one, $h_4$ is initialized as an Identity Mapping Matrix. Only $h_4$ is updated during this stage. Thus all results of MM-GEM \textit{stage one} on cross-modal retrieval and image classification will not be altered.

\noindent\textbf{Evaluation.}
To thoroughly evaluate MM-GEM, we include various downstream tasks in the experiment section. Unless otherwise indicated, all results are reported under the zero-shot protocol without further fine-tuning. We evaluate MM-GEM on: 1) \textbf{Image-Text Retrieval.} For this task, we utilize two prominent benchmark datasets: COCO \cite{lin2014microsoft} and Flickr30K \cite{plummer2015flickr30k}, which feature a diverse collection of images with complex scenes. We evaluate on the standard 1K test set for Flickr30K and 5K test set for COCO. The evaluation metrics are Recall@$K$, where $K=1,5,10$, on both text-to-image and image-to-text retrieval. 2) \textbf{Image Classification.} We evaluate the capability of zero-shot image classification on the track “Image Classification in the Wild” of the ELEVATER
benchmark \cite{li2022elevater}. ELEVATER is designed to challenge models with the task of categorizing images that are captured in real-world, unconstrained environments. The predefined categories of each subset could range from common objects to specific scenes. We follow all metrics of ELEVATER for all subsets. 3) \textbf{Image Captioning.} In this task, we assess the model's ability to generate descriptive and coherent captions for images. We employ two well-established datasets: COCO \cite{lin2014microsoft} and NoCaps \cite{agrawal2019nocaps}. COCO captions include a wide variety of objects, scenes, and activities, while NoCaps encompass novel visual objects. Besides these benchmarks, we also demonstrate MM-GEM's special ability by more evaluation manners. The details will be introduced in the specific sections.

\subsection{Encapsulated Generative and Embedding Objectives}

\subsubsection{Comparison of Training Objectives}
The main risk of MM-GEM is the embedding and generative objectives may conflict with each other in the same text model. The most straightforward way to verify this is to compare the performance of a model trained using the two objectives alone and the objectives together. Therefore, we train three models under the same experimental setting except for training objectives: 1) CLIP-Only which uses embedding objectives $\mathcal{L}_{\text{Emb}}$ alone; 2) Cap-Only which uses generative objectives $\mathcal{L}_{\text{Gen}}$ alone; 3) MM-GEM which uses two objectives simultaneously. These three models adopt ViT-Base as vision encoder and only trained with 1.5 billion seen samples for computational savings. From the result listed in Table \ref{tab:base_retrieval} and \ref{tab:base_cls}, MM-GEM achieves very similar performance to CLIP-Only on both cross-modal retrieval and zero-shot image classification tasks. Table \ref{tab:base_cap} shows that the image captioning performance gap between MM-GEM and Cap-Only is negligible, and this conclusion holds for both stage one and two. For stage one, although MM-GEM lags a little bit behind Cap-Only on COCO in some metrics such as BLEU@4, Rouge and CIDEr, it will be slightly better than Cap-Only on NoCaps, therefore the overall captioning performance is close. After stage-two tuning, the performance gap between Cap-Only and MM-GEM increases. In the meantime we note that MM-GEM still achieves relatively good visual description generation capabilities. Since negligible performance gaps in the first stage can support our conclusions, we leave the continued exploration of this part for future work. These results solidly support our conclusion: encapsulating embedding and generative objectives in the same text model will not lead to significant conflict. 

\begin{table}[t]
  \caption{Performance comparison on zero-shot image captioning task on COCO \cite{lin2014microsoft} and NoCaps \cite{agrawal2019nocaps}. All models in this table use ViT-Base as vision encoder.}
  \label{tab:base_cap}
  \centering
  \resizebox{\linewidth}{!}{
  \begin{tabular}{l cccc cccc}
    \toprule
    \multirow{2}{*}{Model} & \multicolumn{4}{c}{COCO Caption} & \multicolumn{4}{c}{NoCaps Caption}\\
    \cmidrule(lr){2-5}\cmidrule(lr){6-9}
    & BLEU@4  & Meteor  & Rouge & CIDEr & BLEU@4  & Meteor  & Rouge & CIDEr  \\
    \midrule
    Cap-Only \textit{stage one} & 13.4 & 15.8 & 36.2 & 49.8 & 16.6 & 15.9 & 37.3 & 46.3     \\
    MM-GEM \textit{stage one} & 12.9 & 15.8 & 37.0 & 48.8 & 17.3 & 16.2 & 38.6 & 47.0     \\
    Cap-Only \textit{stage two} & 31.2 & 25.4 & 53.2 & 103.9 & 38.8 & 26.6 & 57.0 & 96.5     \\
    MM-GEM \textit{stage two} & 28.7 & 24.5 & 52.0 & 96.3 & 36.2 & 25.7 & 55.6 & 91.0   \\
    \bottomrule
  \end{tabular}
  }
\end{table}

\begin{table}[t]
  \caption{Performance comparison on COCO \cite{lin2014microsoft} and Flickr30K \cite{plummer2015flickr30k} image-to-text (I2T) and text-to-image (T2I) retrieval. R@1, R@5, and R@10 indicate recall ratio at top 1, 5, and 10. All models in this table use ViT-Large as vision encoder.}
  \label{tab:large_retrieval}
  \centering
  \resizebox{\linewidth}{!}{
  \begin{tabular}{l ccc ccc ccc ccc}
    \toprule
    \multirow{2}{*}{Model} & \multicolumn{3}{c}{COCO I2T} & \multicolumn{3}{c}{COCO T2I} & \multicolumn{3}{c}{Flickr30K I2T} & \multicolumn{3}{c}{Flickr30K T2I} \\
    \cmidrule(lr){2-4}\cmidrule(lr){5-7}\cmidrule(lr){8-10}\cmidrule(lr){11-13}
    & R@1  & R@5  & R@10  & R@1  & R@5  & R@10  & R@1  & R@5  & R@10  & R@1  & R@5  & R@10  \\
    \midrule
    CLIP \cite{radford2021learning}     & 57.4 & 80.0 & 87.1 & 34.3 & 58.6 & 69.5 & 87.0 & 97.5 & 99.1 & 63.5 & 86.4 & 91.8    \\
    OpenCLIP \cite{openclip}    & 61.3 & 83.4 & 89.7 & 45.8 & 70.2 & 79.1 & 89.8 & 98.7 & 99.5 & 74.9 & 92.5 & 95.8    \\
    MM-GEM     & 61.0 & 82.6 & 89.2 & 45.6 & 70.5 & 79.3 & 89.0 & 99.0 & 99.5 & 75.4 & 92.6 & 96.0    \\
    
    \bottomrule
  \end{tabular}
  }
\end{table}

\subsubsection{Comparison with State-of-the-arts}
To verify MM-GEM's scalability and make comparison with state-of-the-art works, we train MM-GEM with ViT-Large as vision encoder. For embedding model opponents, we mainly compare with OpenAI CLIP \cite{radford2021learning} and OpenCLIP \cite{openclip}. Table \ref{tab:large_retrieval} shows that MM-GEM achieves similar cross-modal performance to OpenCLIP, while outperforming CLIP by a large margin. This margin may come from the pre-training data: MM-GEM's data is closer to OpenCLIP, while OpenAI CLIP uses private data. In Table \ref{tab:large_cls}, the results are similar to the cross-modal retrieval benchmark. MM-GEM achieves 66.3\% in average, slightly better than OpenCLIP and 4.5\% ahead of OpenAI CLIP. For image captioning models, we choose Flamingo \cite{alayrac2022flamingo} and ClipCap for comparison. Among these models, only ClipCap's training data include COCO. Flamingo bridges contrastive pretrained vision-only models and language-only models by only training a Perceiver Resampler and gated cross attention layers. ClipCap use CLIP encoding as a prefix to the caption, by employing a simple mapping network. From the results in Table \ref{tab:large_caption}, MM-GEM significantly outperforms Flamingo even though the latter adopts a much larger language decoder. As ClipCap is trained on COCO, MM-GEM performs worse than ClipCap on COCO but is substantially ahead on NoCaps.

\begin{table}[t]
  \setlength\tabcolsep{2pt}
  \caption{Performance comparison on Image Classification in the Wild (ICinW) \cite{li2022elevater}. The metric of each dataset follows \cite{li2022elevater}. All models in this table use ViT-Large as vision encoder.}
  \label{tab:large_cls}
  \centering
  \resizebox{\linewidth}{!}{
  \begin{tabular}{l ccccc ccccc ccccc ccccc c}
    \toprule
    Model & \rot{\tiny{CIFAR-10~\cite{cifar}}} & \rot{\tiny{CIFAR-100~\cite{cifar}}} & \rot{\tiny{DTD~\cite{dtd}}} & \rot{\tiny{EuroSAT~\cite{eurosat}}} & \rot{\tiny{FER-2013~\cite{radford2021learning}}} & \rot{\tiny{FGVC-Aircraft~\cite{fgvc}}} & \rot{\tiny{KITTI-Dist.~\cite{kitti}}} & \rot{\tiny{MNIST~\cite{mnist}}} & \rot{\tiny{PatchCamelyon~\cite{patchcamelyon}}} & \rot{\tiny{VOC-2007~\cite{voc}}} & \rot{\tiny{Caltech-101~\cite{caltech}}} & \rot{\tiny{Country-211~\cite{radford2021learning}}} &	\rot{\tiny{Food-101~\cite{food101}}} & \rot{\tiny{GTSRB~\cite{gtsrb}}} &	\rot{\tiny{Hateful Memes~\cite{hateful}}} & \rot{\tiny{Ox. Flowers~\cite{flowers}}} & \rot{\tiny{Ox. IIIT Pets~\cite{pets}}} & \rot{\tiny{Rendered-SST2~\cite{radford2021learning}}} &	\rot{\tiny{RESISC-45~\cite{resisc45}}} &	\rot{\tiny{Stanford-Cars~\cite{stanford_cars}}} & \rot{\tiny{Average}}  \\
    \midrule
    CLIP \cite{radford2021learning}  & 94.0 & 67.4 & 52.6 & 49.5 & 45.5 & 25.7 & 20.5 & 64.4 & 58.4 & 79.5 & 93.0 & 28.1 & 90.2 & 52.9 & 60.2 & 71.4 & 92.2 & 59.9 & 62.3 & 67.4 & 61.8   \\
    OpenCLIP \cite{openclip}  & 96.0 & 82.5 & 61.5 & 65.1 & 47.7 & 32.4 & 22.5 & 65.2 & 57.2 & 80.7 & 94.1 & 25.4 & 89.9 & 56.5 & 54.5 & 74.2 & 92.9 & 60.6 & 72.1 & 91.4 & 66.1   \\
    MM-GEM   & 97.0 & 82.8 & 67.2 & 69.5 & 47.4 & 31.9 & 26.2 & 69.5 & 50.5 & 80.3 & 92.7 & 26.0 & 89.8 & 54.3 & 61.5 & 69.8 & 90.6 & 61.5 & 68.9 & 89.3 & 66.3    \\
    \bottomrule
  \end{tabular}
  }
\end{table}

\begin{table}[t]
  \caption{Performance comparison on image captioning task on COCO \cite{lin2014microsoft} and NoCaps \cite{agrawal2019nocaps}. * denotes model finetuned on COCO training split.}
  \label{tab:large_caption}
  \centering
  \resizebox{\linewidth}{!}{
  \begin{tabular}{l cccc cccc}
    \toprule
    \multirow{2}{*}{Model} & \multicolumn{4}{c}{COCO Caption} & \multicolumn{4}{c}{NoCaps Caption}\\
    \cmidrule(lr){2-5}\cmidrule(lr){6-9}
    & BLEU@4  & Meteor  & Rouge & CIDEr & BLEU@4  & Meteor  & Rouge & CIDEr  \\
    \midrule
    Flamingo-9B \cite{alayrac2022flamingo} & - & - & - & 79.4 & - & - & - & - \\
    ClipCap* & 33.5 & 27.5 & - & 113.1 & - & - & - & 65.8 \\
    MM-GEM  & 32.8 & 26.5 & 54.8 & 110.9 & 39.8 & 27.3 & 57.8 & 100.7     \\
    \bottomrule
  \end{tabular}
  }
\end{table}

\subsection{Fine-grained Understanding Ability}\label{sec:finegrain}
Instead of aligning vision and language modality from the global perspective, MM-GEM is designed to align on the visual feature map. Therefore, MM-GEM is equipped with fine-grained understanding ability. The goal is that PoolAggregator can be directly utilized on the feature map to get the region feature required by embedding or description. However, we observe that regular training of CLIP or image captioning model achieves results that fall far short of this goal. The whole spatial feature map is dominated by the global visual information, making it difficult to distinguish by location. In this paper, we tackle this problem by tuning or adding light projection layers on region-level data, leading to dramatic improvement on fine-grained abilty. In this section, we will present the details in terms of both embedding and description generation.

\subsubsection{Fine-grained Description Generation}
\begin{figure}[t]
  \centering
  \includegraphics[width=0.85\textwidth]{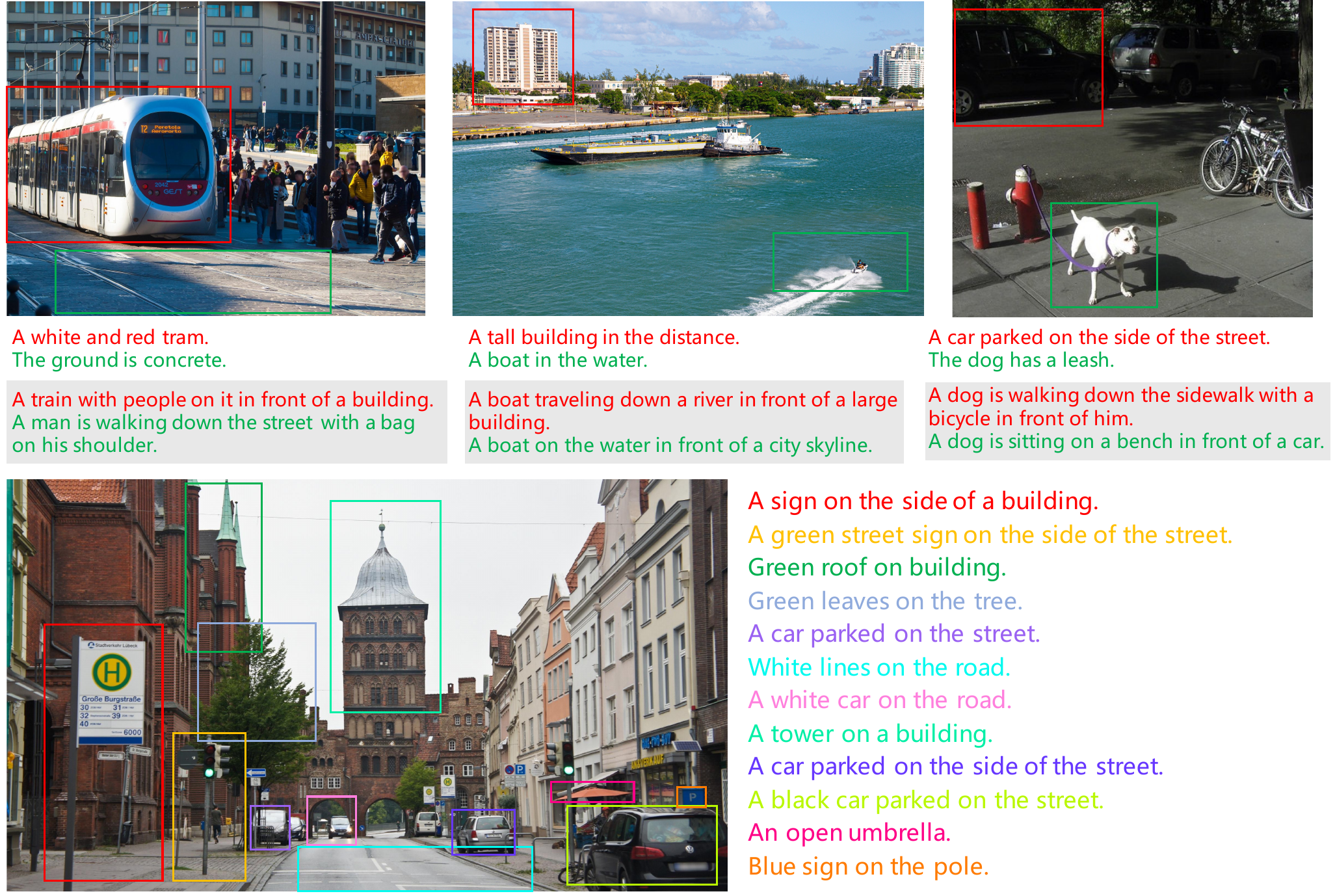}
  \caption{Visualization of fine-grained description generation. This figure shows the captioning results of using region features from the visual feature map as input. The text in the same color as the bounding box in the figure is the description of the corresponding area. Text on a gray background indicates results without region description data.}
  \label{fig:dense_cap}
\end{figure}

A straightforward way to learn image captioning is to flatten a visual feature map and use it as input to a language decoder. The ideal visual feature map has the property that features cropped from a region can generate a description of that region. We find that training on regular image-caption data alone is not sufficient to achieve this property. Therefore, we use the strategy presented in the Section \ref{sec:3.2} to train MM-GEM on mixed data of image-caption and region-description data. We adopt BLIP's filtered data as image-caption data and Visual Genome’s dense caption as region-description data. We compare the results under two different settings and demonstrate in Figure \ref{fig:dense_cap}. The generated region description on gray background shows that the captioning model trained without region-description data fails in distinguishing visual information by location. For example, it tends to generate global image captions or descriptions of non-corresponding locations, like the red bounding box of the building generates ``A boat travelling down a river in front of a large building''. The example in the second row of Figure 2 demonstrates that MM-GEM with region description data can accurately distinguish objects, and still retains the necessary contextual information, like ``A car parked on the side of the street''. These results validate the effectiveness of our stage-two training strategy, avoiding MM-GEM from being limited to a regular image captioning model, and enabling the ability of generating fine-grained descriptions based on regions.

\subsubsection{Fine-grained Image-Text Retrieval}\label{sec:finegrain_retrieval}
Traditional cross-modal embedding models like CLIP focus on global alignment, thus can only act as an instance-level retriever. There are two issues with these models: \textbf{a.} For the text-to-image retrieval task, the query description may only correspond to part of the image. It is worth exploring how well the model can localize the corresponding visual information based on the text query. \textbf{b.} The training approach that focuses only on global alignment may cause the model to focus only on salient objects in the image, which in turn weakens the understanding of fine-grained information. 

\begin{figure}[t]
  \centering
  \includegraphics[width=0.7\textwidth]{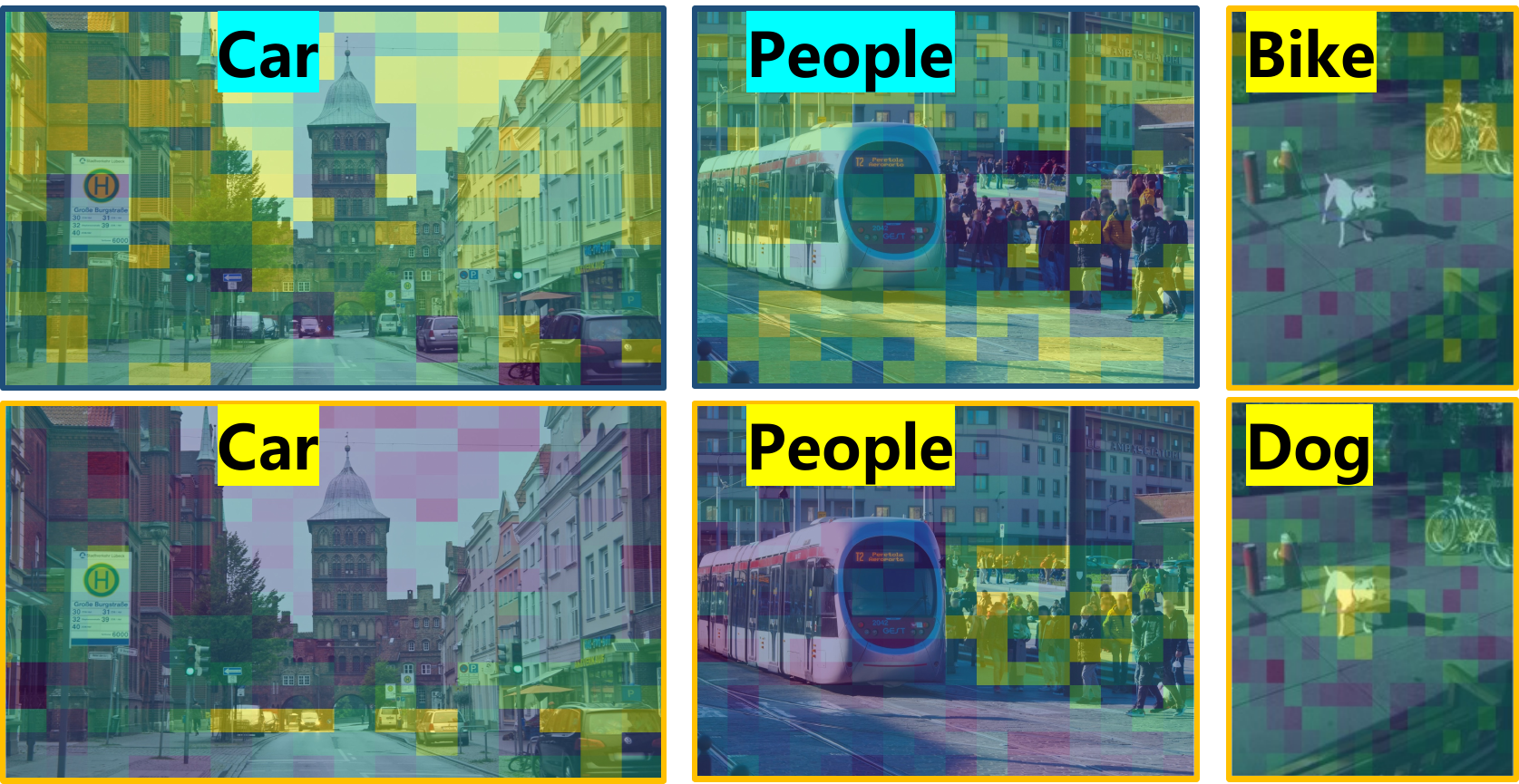}
  \caption{Visualization of fine-grained image-text retrieval. This figure shows the similarity between the visual feature map and the text feature at two stages. The \textcolor{blue}{blue borders and undertones} represent the result from the pre-training stage one, and \textcolor{yellow}{yellow borders and undertones} illustrate the results of stage two. The text superimposed on the image corresponds to the input text.}
  \label{fig:fg_retrieval}
\end{figure}

To study issue \textbf{a.}, we compare MM-GEM trained at two stages by showing the similarities between visual feature maps and several given text queries. For the stage-one model, we calculate the cosine similarities between normalized visual features output by $h_1$ and text features output by $h_2$. For the stage-two model, we alter visual features to the output of $h_4$. The visualized results are demonstrated in Figure \ref{fig:fg_retrieval}. The two columns on the left clearly show that stage-two model well localizes the text query to the corresponding region, while stage-one model totally fails. The rightmost column shows the stage-two model well responds to different text queries for the same image. These results show that stage-two training essentially enables the ability of fine-grained retrieval.

To study issue \textbf{b.}, we add a new quantitative benchmark L-DCI based on Densely Captioned Images (DCI) \cite{urbanek2023picture}. DCI consists of 8012 images from SA-1B \cite{kirillov2023segment}, each image corresponds to a complete description aiming to capture
the full visual details in the image. We directly evaluate cross-modal retrieval on all images and overall descriptions in DCI. We list the text-to-image retrieval results in Table \ref{tab:large_fine_retrieval}. The results in the last two rows show that the stage-two trained MM-GEM improves on all retrieval metrics, especially on L-DCI with nearly 5\% at Recall@1. This result shows that the proposed stage two can improve the model's ability to understand fine-grained information.

\begin{table}[t]
  \caption{Performance comparison on COCO \cite{lin2014microsoft}, Flickr30K \cite{plummer2015flickr30k} and DCI \cite{urbanek2023picture} text-to-image (T2I) retrieval. R@1, R@5, and R@10 indicate recall ratio at top 1, 5, and 10. All models in this table use ViT-Large as vision encoder.}
  \label{tab:large_fine_retrieval}
  \centering
  \resizebox{\linewidth}{!}{
  \begin{tabular}{l ccc ccc ccc}
    \toprule
    \multirow{2}{*}{Model} & \multicolumn{3}{c}{COCO T2I} & \multicolumn{3}{c}{Flickr T2I} & \multicolumn{3}{c}{L-DCI T2I} \\
    \cmidrule(lr){2-4}\cmidrule(lr){5-7}\cmidrule(lr){8-10}
    & R@1  & R@5  & R@10  & R@1  & R@5  & R@10  & R@1  & R@5  & R@10   \\
    \midrule
    CLIP-Only-Base    & 41.7 & 67.2 & 76.7 & 70.3 & 91.4 & 95.3 & 47.3 & 66.3 & 73.0     \\
    MM-GEM-Base   & 41.4 & 66.6 & 76.3 & 70.1 & 90.7 & 95.0 & 46.3 & 66.0 & 72.6     \\
    MM-GEM-Base stage two   & 42.5 & 68.9 & 78.5 & 72.3 & 91.6 & 95.7 & 47.9 & 68.4 & 74.9     \\
    \midrule
    CLIP \cite{radford2021learning}     & 34.3 & 58.6 & 69.5 & 63.5 & 86.4 & 91.8 & 30.4 & 49.4 & 57.3     \\
    OpenCLIP \cite{openclip}    & 45.8 & 70.2 & 79.1 & 74.9 & 92.5 & 95.8 & 43.0 & 62.3 & 69.2     \\
    MM-GEM  & 45.6 & 70.5 & 79.3 & 75.4 & 92.6 & 96.0 & 49.4 & 68.1 & 74.5     \\
    MM-GEM stage two & 47.2 & 72.3 & 81.1 & 76.4 & 94.1 & 96.8 & 54.1 & 72.5 & 78.6     \\
    
    \bottomrule
  \end{tabular}
  }
\end{table}

\subsection{Advanced Text Model in MM-GEM}
As MM-GEM applies a large language model (LLM) as text module, it's critical to figure out other benefits besides introducing generative capabilities. LLMs typically have good ability of language processing due to the large amount of language data. We therefore hypothesize that MM-GEM performs better than regular CLIP on data containing more complex text. To verify this, we mainly focus on long-form text image caption data in this work. We evaluate MM-GEM and regular CLIP on L-DCI cross-modal retrieval benchmark described in Section \ref{sec:finegrain_retrieval}. Results shown in Table \ref{tab:large_dci} indicate that an advanced text module will significantly improves the performance on the benchmark with long-form text, by over 5\% margin on Recall@1. And the performance further increases while the image size is 336. It is worth noting that MM-GEM was trained with a maximum text length of 50, which is shorter than CLIP. We adjusted it to 200 when testing on DCI only. The comparison between CLIP-Only-Base and MM-GEM-Base demonstrate that the improvement comes from the text module instead of training objectives. Even though MM-GEM does not show advantage on typical cross-modal retrieval benchmarks like COCO and Flickr30K, according to Table \ref{tab:large_retrieval}, the improvement on a more complex benchmark is significant. The results in this section inspire future works on exploring benefits of an advanced text encoder in CLIP.

\begin{table}[t]
  \caption{Long-form text image retrieval performance comparison on DCI \cite{urbanek2023picture}. R@1, R@5, and R@10 indicate recall ratio at top 1, 5, and 10. }
  \label{tab:large_dci}
  \centering
  \begin{tabular}{l ccc ccc}
    \toprule
    \multirow{2}{*}{Model} & \multicolumn{3}{c}{L-DCI I2T} & \multicolumn{3}{c}{L-DCI T2I} \\
    \cmidrule(lr){2-4}\cmidrule(lr){5-7}
    & R@1  & R@5  & R@10  & R@1  & R@5  & R@10  \\
    \midrule
    CLIP-Only-Base     & 46.7 & 67.4 & 73.9 & 47.3 & 66.3 & 73.0 \\
    MM-GEM-Base & 46.6 & 67.4 & 74.4 & 46.3 & 66.0 & 72.6 \\
    \midrule
    CLIP-Large \cite{radford2021learning}    & 33.7 & 52.8 & 60.0 & 30.4 & 49.4 & 57.3    \\
    OpenCLIP-Large \cite{openclip}    & 44.4 & 64.0 & 70.5 & 43.0 & 62.3 & 69.2    \\
    MM-GEM-Large     & 49.4 & 69.4 & 75.4 & 49.4 & 68.1 & 74.5 \\
    MM-GEM-Large-336     & 51.0 & 70.2 & 76.5 & 51.8 & 70.2 & 76.0     \\
    
    \bottomrule
  \end{tabular}
\end{table}

%% file: sections/05_conclusion.tex
\section{Conclusion}
The Multi-Modal Generative Embedding Model (MM-GEM) presents a unified approach to multi-modal learning by integrating generative and embedding objectives within a single Large Language Model (LLM). Our experiments demonstrate that these two objectives do not significantly
conflict with each other. MM-GEM achieves competitive performance across a range of tasks, including cross-modal retrieval, zero-shot classification, and image captioning. A key contribution is the PoolAggregator, enhancing the model's ability to handle fine-grained tasks. Additionally, MM-GEM's advanced text module significantly improves performance on long-form text retrieval, showcasing the benefits of leveraging a robust LLM for text processing.

MM-GEM represents a significant step towards unified multi-modal models, yet there are still many subsequent potential directions: 1) We mainly focus on image captioning for generative tasks in this work, the performance impact of adding plain language data needs to be further investigated. 2) MM-GEM enables LLMs generate discriminative outputs besides language tokens, this may benefit multi-modal large language model by retrieving or grounding visual information efficiently. Further investigation into these aspects will be explored in future work.

\section{Limitations}
\label{sec:limitations}
Although MM-GEM integrates generative and embedding objectives within a single LLM, there are still some limitations. 
Caption loss allows the model to focus more on the detailed information in the text, but the existing dataset limits its ability. The text we used in pre-training stage one is noisy, of low quality, and has less detailed information, which limits the ability of our model. 
The design of the PoolAggregator allows our model to handle fine-grained tasks. However, in pre-training stage one, the model does not directly exhibit fine-grained capabilities due to the lack of region-level data. Still, the implied fine-grained capabilities require only a very small amount of region-level data to be bootstrapped. We can try to introduce a very small amount of region-level data in pre-training stage one as well, so that the model can have a stronger fine-grained capability.